\documentclass[conference]{IEEEtran}
\IEEEoverridecommandlockouts
\usepackage{cite}
\usepackage{amsmath,amssymb,amsfonts}
\usepackage{algorithmic}
\usepackage{graphicx}
\usepackage{textcomp}
\usepackage{xcolor}
\usepackage{multirow}
\usepackage{booktabs}
\usepackage{soul}
\usepackage{comment}
\usepackage{array}
\usepackage{tabularx}
\def\BibTeX{{\rm B\kern-.05em{\sc i\kern-.025em b}\kern-.08em
    T\kern-.1667em\lower.7ex\hbox{E}\kern-.125emX}}
\begin{document}

\title{CST: Collaborative Selective Transmission for Communication-Efficient Multimodal Edge Inference%
\thanks{This work has been submitted to the IEEE for possible publication.
Copyright may be transferred without notice, after which this version may no longer be accessible.}}

\author{
\IEEEauthorblockN{
Hai Chi\IEEEauthorrefmark{1},
Junrui Zhang\IEEEauthorrefmark{2},
Rui Ning\IEEEauthorrefmark{2},
Chonggang Wang\IEEEauthorrefmark{3},
Robert Gazda\IEEEauthorrefmark{3},
Huanrui Yang\IEEEauthorrefmark{1},
Hongyi Wu\IEEEauthorrefmark{1}
}

\IEEEauthorblockA{
\IEEEauthorrefmark{1}
Department of Electrical and Computer Engineering,
University of Arizona, Tucson, AZ, USA\\
\{chi149, huanruiyang, mhwu\}@arizona.edu
}

\IEEEauthorblockA{
\IEEEauthorrefmark{2}
Department of Computer Science,
Old Dominion University, Norfolk, VA, USA\\
jzhan008@odu.edu, rning@cs.odu.edu
}

\IEEEauthorblockA{
\IEEEauthorrefmark{3}
InterDigital, Conshohocken, PA, USA\\
\{Chonggang.Wang, robert.gazda\}@interdigital.com
}
}

\maketitle

\begin{abstract}
Collaborative multimodal inference improves edge perception by combining observations from distributed sensing devices, but transmitting high-dimensional helper representations incurs substantial communication overhead and can lead to high end-to-end latency. Existing
communication-efficient methods reduce payloads through
compression, semantic coding, or feature selection, yet typically
optimize compactness or task relevance without explicitly accounting
for information already represented at the main device. Consequently,
task-relevant but redundant helper features may still consume
bandwidth. We present Collaborative Selective Transmission (CST), a
main-directed query--response framework that retrieves only helper
information complementary to the current main representation. Inspired
by Partial Information Decomposition and the Multiview Redundancy
Assumption, CST learns sample-adaptive, helper-specific sparse
retrieval supports while discouraging retrieval of semantics already covered by the
main device or duplicated across helpers. During inference, the main
device transmits only support indices, and each helper returns the
corresponding latent values, avoiding dense helper-feature exchange.
Across three real-world multimodal sensing benchmarks, CST transmits no more than 14.18\% of helper feature values while achieving best or near-best
task performance among the evaluated methods. Experiments on a five-node NVIDIA Jetson Orin Nano testbed across 5--100\,Mbps demonstrate up to a $4.27\times$ speedup over Transmit-All in end-to-end inference, confirming practical end-to-end latency reductions.
\end{abstract}

\begin{IEEEkeywords}
collaborative inference, multimodal sensing, selective transmission, partial information decomposition
\end{IEEEkeywords}

\section{Introduction}
\label{sec:introduction}

Real-time edge applications increasingly rely on multimodal sensing to
support accurate and responsive perception in ambient monitoring, human activity understanding, and autonomous
systems~\cite{IoTSurvey,EdgeComputingSurvey,endedgecloud,
InfoFusionandEdgeComputingSurvey}. A single edge device, however, is constrained by its local sensing
coverage, available modalities, and on-device computing resources.
Collaborative edge inference addresses these limitations by allowing
the device responsible for the target task, referred to as the main
device, to integrate synchronized observations collected by
spatially distributed helper devices across heterogeneous sensing
modalities~\cite{multimodalsensorfusionauto,
collaborativeperceptionforautonomousdriving}. Exchanging intermediate representations rather than raw sensor streams
keeps sensing and feature extraction local while allowing the main
device to exploit complementary evidence and improve task performance~\cite{splitcomputingearlyexiting}.

\begin{figure*}[t]
    \centering
    \includegraphics[width=\textwidth]{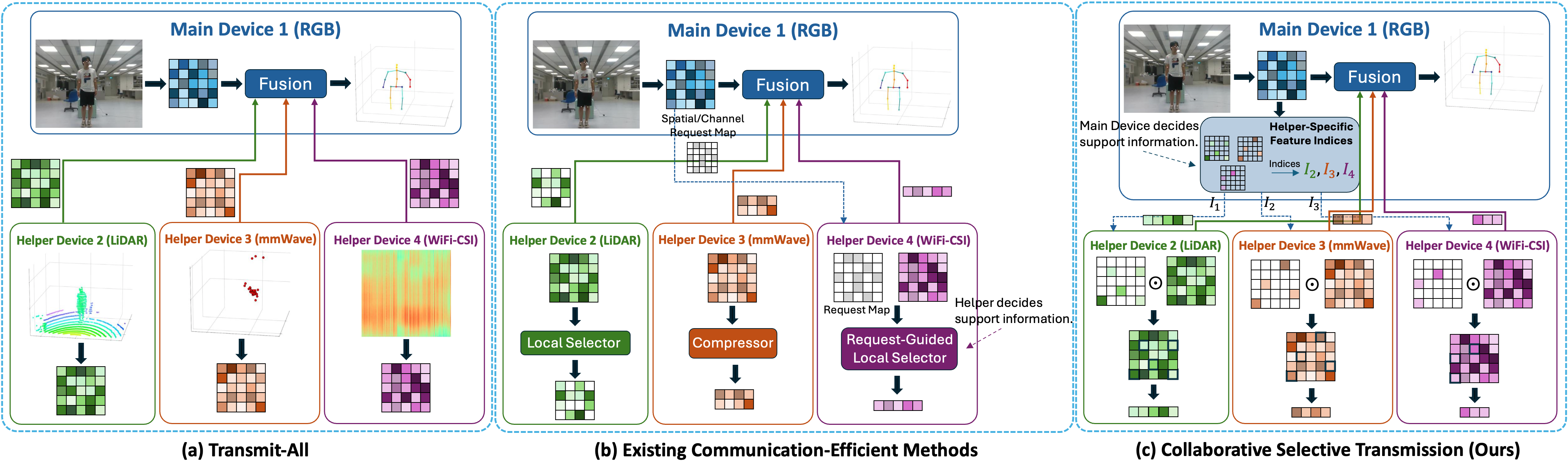}
    \caption{Collaborative multimodal transmission with three representative
    MM-Fi~\cite{mmfi} helpers in a 3D pose estimation example.
    (a) Transmit-All sends complete helper representations.
    (b) Representative methods use local selection, learned compression,
    or request-guided helper-side selection; the paths denote distinct
    method families.
    (c) CST generates helper-specific sparse supports from the current main
    representation and retrieves the corresponding values.}
    \label{fig:intro}
\end{figure*}

Collaboration, however, can shift a substantial fraction of the inference
cost to network communication. The most direct strategy is Transmit-All, illustrated in
Figure~\ref{fig:intro}(a), which transmits the complete intermediate representation of every
helper. Although this strategy avoids
discarding potentially useful features, the aggregate payload grows with both the
number of helpers and the dimensionality of their representations.
Moreover, many transmitted features may encode task-relevant semantics
already represented at the main device. The resulting redundant traffic accumulates across helpers and can
make network transmission the dominant component of end-to-end inference latency.

Communication-efficient methods reduce this cost through learned
compression, information-bottleneck coding, semantic transmission, and
dynamic feature selection~\cite{bottlenetpp,vibdyn,vddibsr,sinfony,dtjscc,dynamicvit}. Some collaborative
methods additionally use receiver-generated requests: Where2Comm
targets low-confidence spatial regions, and How2Comm coordinates
spatial--channel selection~\cite{where2comm,how2comm}. These requests
specify structured regions or channels, and the final transmitted
subset still depends on helper-side confidence or attention. Moreover, these selection mechanisms are not trained against an explicit redundancy reference derived from the current main representation. Consequently, these methods may still transmit helper information already covered by the main representation, leaving avoidable communication overhead.

\begin{figure}[!htb]
\centering
\includegraphics[width=\columnwidth]{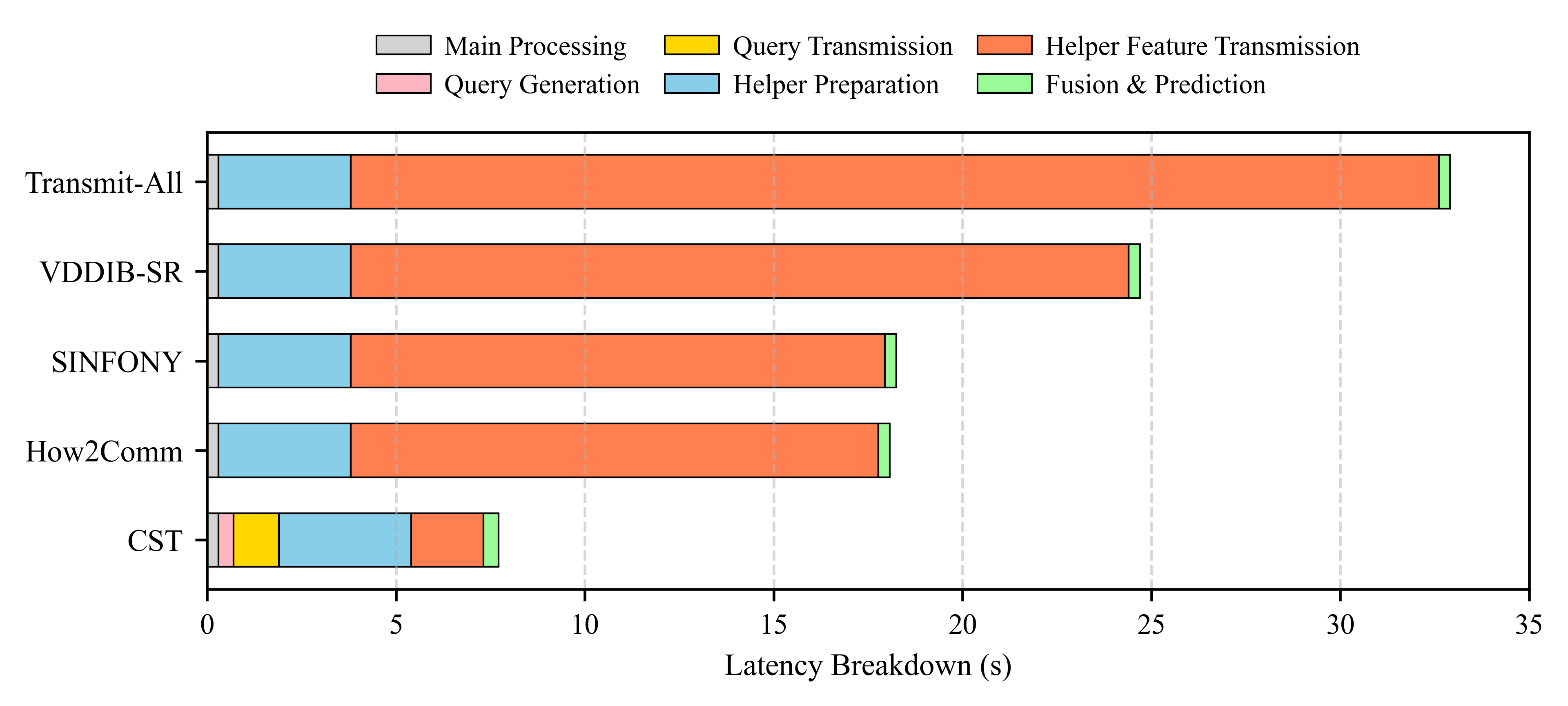}
    \caption{Latency profiling on MM-Fi~\cite{mmfi} in a network under a shared aggregate bandwidth of 5 Mbps, with RTT and jitter configured to 30 ms and 5 ms, respectively, and a batch size of 32. Communication dominates the end-to-end latency in all shown baseline methods, including Transmit-All, VDDIB-SR~\cite{vddibsr}, SINFONY~\cite{sinfony}, and How2Comm~\cite{how2comm}, whereas CST achieves a total latency of 7.71\,s.}
    \label{fig:intro_latency_breakdown}
\end{figure}

To quantify the communication bottleneck that persists despite these optimizations, we deploy the five MM-Fi~\cite{mmfi} sensing modalities on five active NVIDIA Jetson Orin Nano nodes, with one modality per node. One node serves as the main device, and the remaining four serve as helpers. The testbed and experimental setup are described in Section~\ref{sec:evaluation}. Under a shared aggregate bandwidth of 5\,Mbps, configured round-trip time (RTT) and jitter targets of 30\,ms and 5\,ms, respectively, and a batch size of 32, Figure~\ref{fig:intro_latency_breakdown} shows that query and
helper-feature transmission together account for at least 75\% of the
batch-level end-to-end latency across all baselines shown. These results demonstrate that, even after existing communication optimizations, network transmission remains the dominant end-to-end latency component and the greatest opportunity for further end-to-end acceleration.

To address this limitation, we propose Collaborative Selective
Transmission (CST), shown in Figure~\ref{fig:intro}(c). Rather than merely reducing or filtering each helper representation,
CST learns to retrieve only those helper features that provide
task-relevant information missing or insufficiently represented in the
current main representation. 
The main device directly generates a distinct sparse retrieval support for each helper, sends the corresponding indices, and receives only the requested latent values. As shown in Figure~\ref{fig:intro_latency_breakdown}, even after accounting for all query and coordination overhead, CST reduces the batch-level end-to-end latency to 7.71\,s, substantially outperforming all plotted baseline methods.

CST is inspired by Partial Information Decomposition
(PID)~\cite{quantifyuniqueinfo,demopositionMI} and the Multiview
Redundancy Assumption~\cite{CoMM,FactorCL}. PID motivates separating
information already covered by the main device from complementary
helper information. The Multiview Redundancy Assumption further
suggests that, after conditioning on the main view, the additional
task-relevant information contributed by a helper view is limited.
Thus, preserving the entire high-dimensional helper representation may
be unnecessary. This motivates identifying and
transmitting a compact subset of helper features that retains
task-relevant information not sufficiently captured by the main
representation. CST does not explicitly compute PID components during training.
Instead, it realizes this functional separation through tractable
end-to-end objectives that discourage the retrieval of main-covered or
duplicated semantics while preserving complementary task information.

This paper makes the following contributions:

\begin{itemize}
    \item \textbf{PID-motivated communication objective.}
    We formulate efficient helper communication in terms of the
    task-relevant information gain beyond the current main representation,
    rather than the complete helper representation. PID and the Multiview
    Redundancy Assumption motivate separating main-covered semantics from
    complementary helper information.

    \item \textbf{Main-directed sparse feature retrieval.}
    We design CST as a main-directed query--response protocol in which the
    current main representation is used to generate a distinct sparse
    retrieval support for each heterogeneous helper. The main device
    transmits only the support indices, and each helper returns the
    corresponding latent values without executing a helper-side selection
    module. A training-time redundancy reference further encourages the learned
    supports to identify helper features carrying task-relevant information
    that is missing or insufficiently represented in the main
    representation.

    \item \textbf{Comprehensive task and system evaluation.}
    We evaluate CST on three real-world multimodal sensing benchmarks
    against Transmit-All and multiple communication-efficient
    methods. CST transmits at most 14.18\% of the available helper feature
    values while achieving best or near-best task performance among the evaluated methods. Experiments on a physical multi-device edge testbed further demonstrate latency reductions across shared aggregate bandwidths of 5--100\,Mbps, with up to a $4.27\times$ end-to-end speedup.
\end{itemize}

\section{Problem Formulation and Theoretical Insight} 
\label{sec:motivation}
In this section, we formalize the collaborative edge inference problem and provide theoretical insights into information redundancy to motivate our system design.

\subsection{Problem Formulation}
Following standard collaborative perception settings~\cite{edgemultiview}, given one main device and a set of synchronized helper devices, our goal is to extract a compact feature subset that
enhances the main device's task performance while matching or
surpassing traditional full-fusion methods. For a given set of $n$ devices $\mathcal{D} =\{\mathcal{D}_1, \mathcal{D}_2, \dots, \mathcal{D}_n\}$, the set of inputs is denoted as $X=\{X_1, X_2, \dots, X_n\}$, where $X_i \in \mathbb{R}^{T_i \times d_i}$. Here $T_i$ denotes the input length of $\mathcal{D}_i$, and $d_i$
denotes the input dimension at each position. Let $F_i$ denote the intermediate feature representation of device $\mathcal{D}_i$, extracted by a local encoder $\mathcal{E}_i(\cdot)$ such that $F_i = \mathcal{E}_i(X_i)$. The dimensionality of $F_i$ may differ across devices due to heterogeneous modalities. Without loss of generality, let $\mathcal{D}_1$ serve as the main device and $\{\mathcal{D}_i\}_{i=2}^n$ act as helper devices. Our objective is to identify a feature subset $\hat{F}_i$ (denoted as $Z_c^i$ in Section~\ref{sec:complementary_selection}) for each helper $i$, minimizing the helper feature-value payload $\sum_{i=2}^n \|\hat{F}_i\|_0$
(where $\|\cdot\|_0$ denotes the $\ell_0$ pseudo-norm, i.e., the
number of nonzero transmitted feature values) while maximizing the main device's task performance.

\subsection{Theoretical Insight: Partial Information Decomposition}
\label{sec:pid}
Our system design is motivated by \emph{Partial Information Decomposition}
(PID)~\cite{quantifyuniqueinfo, demopositionMI}, an information-theoretic framework that characterizes
how multiple information sources jointly contribute to a target variable.
Unlike conventional mutual information, PID conceptually decomposes the task-relevant information from
multiple sources into three types of components: \emph{unique information}, which is
available from only one source; \emph{redundant information}, which is shared
across sources; and \emph{synergistic information}, which can only be recovered
through jointly combining multiple sources.

For a collaborative system consisting of a main device with feature
representation $F_1$ and a helper device with representation $F_j$, PID
expresses the joint information about the prediction target $Y$ as
\begin{equation}
I(F_1,F_j;Y)
=
U_1
+
U_j
+
R_{1j}
+
S_{1j},
\label{eq:pid}
\end{equation}
where $U_1$ and $U_j$ denote the unique information contributed by each
device, $R_{1j}$ represents redundant information already shared between
them, and $S_{1j}$ denotes synergistic information that emerges only
through joint reasoning.

This decomposition provides an important theoretical insight for collaborative
edge inference. In the PID interpretation, $R_{1j}$ denotes task-relevant information
already available from the main representation; transmitting it
therefore does not increase the task information available at the main
device. Instead, communication should ideally convey only the information unavailable
at the main device, namely the helper's unique information together with the
portion of synergistic information necessary for collaborative reasoning.
To this end, PID naturally defines the desired objective of communication:
maximize the transmission of complementary information while minimizing
redundant information.

This intuition is further supported by the widely adopted
\emph{Multiview Redundancy Assumption} in multimodal representation
learning~\cite{CoMM,FactorCL,
tsai2021selfsupervisedlearningmultiviewperspective}, which assumes that,
for a small constant $\epsilon>0$,
\begin{equation}
I(Y;F_j\mid F_1)<\epsilon.
\label{eq:bound}
\end{equation}
For a main--helper pair $(F_1,F_j)$, this conditional gain consists of the helper's unique information and
the helper-side evidence required to form useful synergy with the main
representation.

Eq.~(\ref{eq:bound}) suggests that, for correlated modalities, the
task-relevant gain from a helper after observing the main
representation may remain small even when its feature representation
is high-dimensional. Although this assumption does not imply
coordinate-wise sparsity, it motivates seeking a compact subset
that preserves the remaining conditional task information.

Because explicitly estimating PID components during CST training is
impractical~\cite{CoMM,liang2023quantifyingmodelingmultimodal},
CST realizes the PID-motivated redundancy and complementarity roles
through lightweight anchors and tractable end-to-end objectives, as
described in Section~\ref{sec:method}.

\section{Proposed CST Framework}
\label{sec:method}

\begin{figure*}[t]
  \centering
  \includegraphics[width=\textwidth]{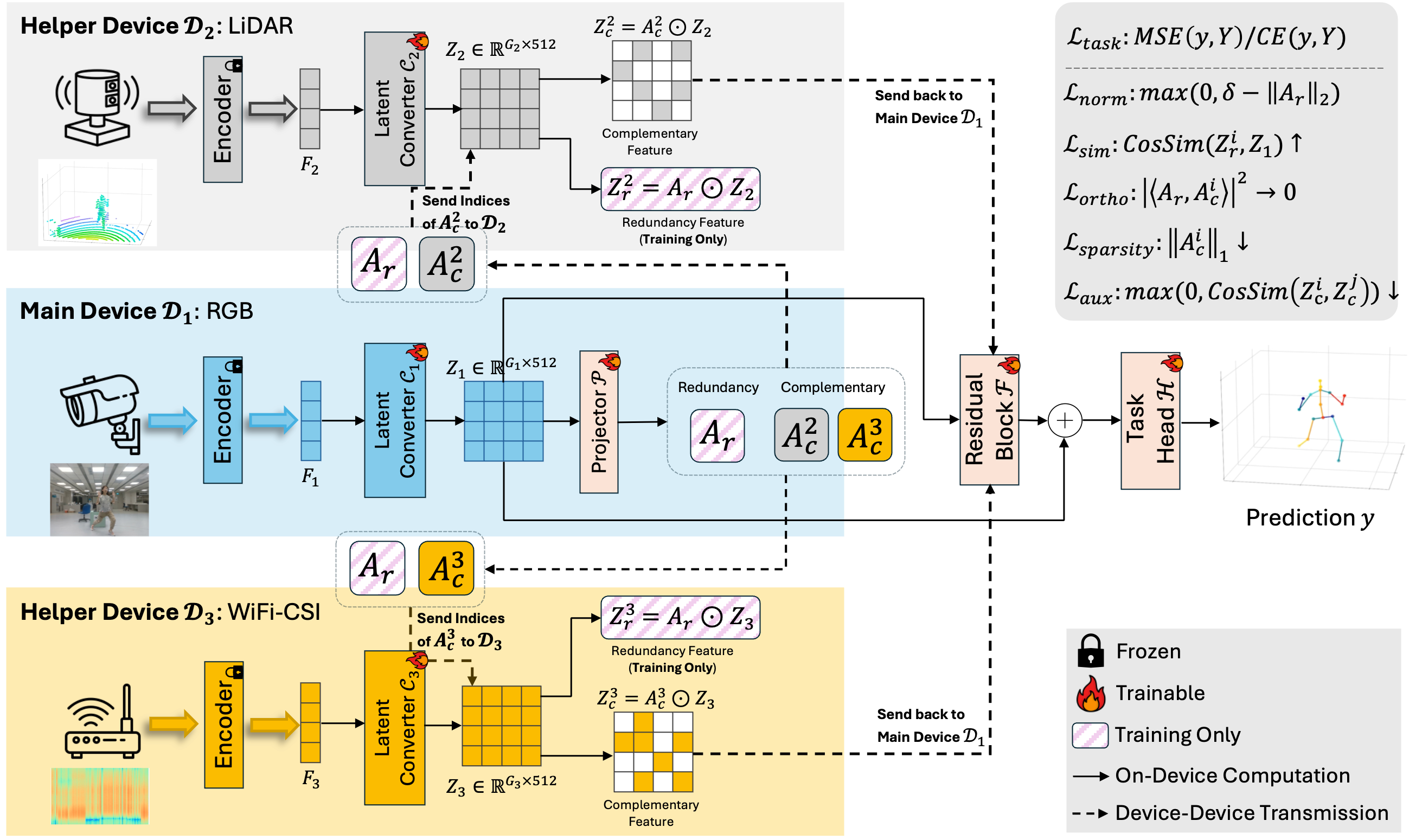} 
  \caption{Overview of CST, illustrated with an MM-Fi 3D pose estimation
example using RGB as the main modality and LiDAR and WiFi-CSI as two
representative helpers. The main representation generates a shared
training-time redundancy anchor and helper-specific complementary
anchors. During inference, the main device sends only the support
indices of each complementary anchor, and the helpers return the
requested latent values for residual fusion and task prediction.
Additional helpers follow the same query--response path. }
  \label{fig:system_overview}
\end{figure*}

As illustrated in Figure~\ref{fig:system_overview}, CST consists of a
standard feature preparation stage followed by its selective
transmission mechanism. Each device first obtains a modality-specific
representation and maps it into a common latent space. CST then uses
the main-device representation to distinguish information already
available locally from helper evidence that can improve the task
prediction. Only the selected helper features are transmitted to the
main device for collaborative prediction.

\subsection{Feature Preparation}
\label{sec:framework_overview}

As in standard collaborative multimodal inference pipelines~\cite{edgemultiview}, each
device first performs local feature extraction and preparation before
collaborative communication and inference.
Let $\mathcal{D}_1$ denote the main device and
$\{\mathcal{D}_i\}_{i=2}^{n}$ the helper devices. Following the common practice of modality-specific feature extraction and lightweight latent projection~\cite{CoMM,FactorCL}, the main device
applies a pretrained encoder $\mathcal{E}_1$ followed by a trainable
latent converter $\mathcal{C}_1(\cdot)$, parameterized by
$\theta_c^1$. Similarly, helper $\mathcal{D}_i$ applies
$\mathcal{E}_i$ followed by $\mathcal{C}_i(\cdot)$, parameterized by
$\theta_c^i$:
\begin{equation}
    F_i=\mathcal{E}_i(X_i), \qquad
    Z_i=\mathcal{C}_i(F_i), \quad 1\leq i\leq n.
    \label{eq:feature_preparation}
\end{equation}
Here, $X_1$ and $X_i$ are the local observations at the main device
and helper $\mathcal{D}_i$, respectively. $F_1$ and $F_i$ are the
corresponding modality-specific intermediate features, while $Z_1$ and
$Z_i$ are their representations in the common latent space.
Accordingly, $Z_1$ is available at the main device, while
$Z_i$, $2\leq i\leq n$, is maintained locally by helper
$\mathcal{D}_i$.

Each aligned representation satisfies
$Z_i\in\mathbb{R}^{G_i\times512}$, where $G_i$ denotes the modality-specific latent sequence
length, e.g., the number of spatiotemporal grid locations in RGB or
thermal features, or encoded feature locations in LiDAR representations. The modality-specific encoders remain frozen throughout CST training and inference, whereas the latent converters are optimized during CST
training. The main representation $Z_1$ is used for anchor generation and final
fusion, while each helper representation $Z_i$ provides the feature
values eligible for retrieval.

\subsection{Complementary Feature Selection}
\label{sec:complementary_selection}

The central challenge is to determine what task-relevant information
is still missing from the main representation. Rather than treating each helper representation
$Z_i$ as an indivisible message, CST seeks the task-relevant evidence
that becomes useful beyond the main representation $Z_1$. From the PID
perspective in Section~\ref{sec:pid}, this evidence includes information
unique to the helper and the helper-side information required to form
useful synergy with $Z_1$. Under the Multiview Redundancy Assumption,
Eq.~(\ref{eq:bound}) states that the task-relevant conditional gain
available from a helper after observing the main representation is bounded by a constant $\epsilon$. This bound motivates CST to represent the helper's conditional gain with a compact complementary feature.

To identify this complementary evidence, CST employs a trainable
projector $\mathcal{P}$, parameterized by $\theta_p$, which takes only
$Z_1$ as input and generates a shared redundancy anchor $A_r$ and a
helper-specific complementary anchor $A_c^i$ for each helper:
\begin{equation}
    [A_r;\{A_c^i\}_{i=2}^{n}]
    =
    \mathcal{P}(Z_1;\theta_p),
    \label{eq:projector}
\end{equation}
where $[\cdot;\cdot]$ denotes the collection of projector outputs.
ReLU is applied only to $A_c^i$, whose strictly positive entries
define the scalar-level transmission support. No activation is applied
to $A_r$, which remains a continuous training-time reference for
main-covered helper semantics. When $G_i\neq G_1$, $Z_1$ is linearly resampled to $G_i$ before
generating $A_c^i$, and the shared $A_r$ token map is resampled to
$G_i$ before helper-specific operations. In the following helper-specific expressions, $Z_1$ and $A_r$ refer
to their versions aligned to $G_i$. This operation aligns only the token-sequence length and does not
assume token-wise spatial correspondence across modalities;
helper-specific projector heads learn the cross-modal conditioning end
to end. Each $A_c^i$ is dynamically generated for the current sample and
helper, rather than learned as a fixed gating tensor. The complementary anchor is applied element-wise to the corresponding helper representation:
\begin{equation}
    Z_c^i
    =
    Z_i\odot A_c^i,
    \label{eq:complementary_feature}
\end{equation}
where $\odot$ denotes Hadamard product. The resulting $Z_c^i$
represents the complementary helper contribution used together with
$Z_1$ for prediction.

\subsection{End-to-End Complementary Anchor Learning}
\label{sec:complementary_learning}

Section~\ref{sec:complementary_selection} specifies the desired role of
$A_c^i$: retaining the task-relevant contribution of helper
$\mathcal{D}_i$ that is not sufficiently represented in the main
representation $Z_1$.
Under the pairwise PID interpretation in Section~\ref{sec:pid}, the additional
task-relevant contribution of helper $\mathcal{D}_i$ beyond the main
representation conceptually corresponds to $U_i+S_{1i}$, whereas
$R_{1i}$ denotes task-relevant information shared by the main and
helper views and is therefore already available at the main device.
Here, $U_i$ is unique to the helper, while $S_{1i}$ emerges only when
the main and helper representations are used jointly. CST therefore seeks to retain helper-unique evidence and the helper-side
features needed to realize useful main--helper synergy, while avoiding
the retrieval of semantics already represented in $Z_1$. This interpretation specifies
the desired behavior of $A_c^i$, but does not directly provide a
tractable supervision signal for learning it.

Direct end-to-end selection leaves the distinction between main-covered
and complementary helper components implicit. CST therefore introduces
$A_r$ as a training-time reference for helper semantics already covered
by the main representation. This explicit redundancy reference guides
each helper-specific $A_c^i$ to focus on information not already
represented by $Z_1$. An ablation study is presented in Section~\ref{sec:ablation} to evaluate the
necessity of this design.

To construct a training-time reference for helper semantics already
represented in $Z_1$, CST applies the shared redundancy anchor $A_r$
to $Z_i$:

\begin{equation}
    Z_r^i
    =
    Z_i\odot A_r,
    \label{eq:redundancy_feature}
\end{equation}
where $Z_r^i$ is a training-only proxy for helper semantics also
represented in $Z_1$. The redundancy-anchor generator is shared,
whereas its aligned output and the resulting $Z_r^i$ are
helper-specific. Unlike $A_c^i$, $A_r$ does not define a communication
support. It is used only during training to provide supervision for
complementary feature learning.

Together, $Z_r^i$ and $Z_c^i$ implement the two functional roles
motivated above. The redundancy branch $Z_r^i$ is trained to capture
helper semantics already covered by $Z_1$, whereas the complementary
branch $Z_c^i$ is optimized to retain task-relevant evidence beyond
$Z_1$. As discussed in Section~\ref{sec:pid}, directly estimating PID components from
high-dimensional neural representations during CST training is
impractical. CST enforces these roles through tractable loss
functions~\cite{liang2023quantifyingmodelingmultimodal}.

To encourage $Z_r^i$ to capture helper semantics already represented
at the main device, CST aligns it with $Z_1$. Cosine similarity is
computed after mean pooling each sequence representation over the token
dimension:
\begin{equation}
    \mathcal{L}_{\mathrm{sim}}
    =
    \frac{1}{n-1}
    \sum_{i=2}^{n}
    \left(
        1-\operatorname{CosSim}(Z_r^i,Z_1)
    \right).
    \label{eq:Lsim}
\end{equation}
This objective encourages $A_r$ to emphasize helper feature components
that are semantically aligned with the main representation. Because
cosine alignment does not constrain the magnitude of $A_r$, and the
separation objective admits the trivial solution
$A_r\rightarrow0$, we prevent this degeneration using
\begin{equation}
    \mathcal{L}_{\mathrm{norm}}
    =
    \max\left(
        0,
        \delta-\|A_r\|_2
    \right),
    \label{eq:LossNorm}
\end{equation}
where $\delta>0$ is a positive margin that maintains a nontrivial
redundancy reference.

Once this reference is established, each complementary anchor is
explicitly separated from it:
\begin{equation}
    \mathcal{L}_{\mathrm{ortho}}
    =
    \frac{1}{n-1}
    \sum_{i=2}^{n}
    \left|
        \left\langle A_r,A_c^i\right\rangle
    \right|^2.
    \label{eq:LossOrtho}
\end{equation}

Here, $\langle\cdot,\cdot\rangle$ denotes the feature-wise inner
product averaged over the batch and token positions. Minimizing $\mathcal{L}_{\mathrm{ortho}}$ discourages $A_c^i$ from
emphasizing the same feature components as the redundancy reference
$A_r$, encouraging it to focus on information beyond the current main
representation.

Eq.~(\ref{eq:bound}) suggests that the additional task-relevant
information available from a helper after observing the main view may
be limited. Although this information-theoretic bound does not directly
imply coordinate-wise sparsity, it motivates seeking a compact helper
subset that preserves the remaining task-relevant gain. To realize this
objective, CST encourages each complementary anchor to use a compact
support. Since the $\ell_0$ communication objective in
Section~\ref{sec:motivation} is non-differentiable, we apply an
$\ell_1$ regularizer directly to the ReLU-activated complementary
anchors:
\begin{equation}
    \mathcal{L}_{\mathrm{sparsity}}
    =
    \frac{1}{n-1}
    \sum_{i=2}^{n}
    \|A_c^i\|_1.
    \label{eq:Lsparsity}
\end{equation}
Because the strictly positive entries of $A_c^i$ define the
scalar-level transmission support, this regularizer encourages
unnecessary activations to become zero, thereby reducing the number of
transmitted helper values.

When multiple helpers are present, independently learned
complementary representations may still contain overlapping evidence.
For $n>2$, CST introduces
\begin{equation}
    \mathcal{L}_{\mathrm{aux}}
    =
    \frac{2}{(n-1)(n-2)}
    \sum_{2\leq i<j\leq n}
    \max\left(
        0,
        \operatorname{CosSim}(Z_c^i,Z_c^j)
    \right),
    \label{eq:Laux}
\end{equation}
which discourages positively aligned complementary contributions from
different helpers. When $n=2$, no helper--helper pair exists and
$\mathcal{L}_{\mathrm{aux}}$ is set to zero.

The learned complementary representations are then combined with the
main representation through a trainable fusion module $\mathcal{F}$,
parameterized by $\theta_f$:
\begin{equation}
    Z_{\mathrm{final}}
    =
    Z_1+
    \mathcal{F}
    \left(
        Z_1,\{Z_c^i\}_{i=2}^{n};
        \theta_f
    \right).
\end{equation}
The fusion module jointly processes $Z_1$ and the complementary
representations $\{Z_c^i\}_{i=2}^{n}$ to produce a task-relevant
residual update. The outer skip connection preserves $Z_1$, allowing
helper evidence to augment rather than replace the main
representation~\cite{resnet}. A trainable task head $\mathcal{H}$, parameterized by
$\theta_h$, maps $Z_{\mathrm{final}}$ to the prediction $\mathbf{y}$.

The task loss encourages the sparse features retained by $A_c^i$ to
remain useful when combined with $Z_1$:
\begin{equation}
    \mathcal{L}_{\mathrm{task}}
    =
    \begin{cases}
        \displaystyle
        -\frac{1}{N_b}
        \sum_{k=1}^{N_b}
        \mathbf{Y}_k\cdot\log\mathbf{y}_k,
        & \text{classification},\\[12pt]
        \displaystyle
        \frac{1}{N_b d_y}
        \sum_{k=1}^{N_b}
        \left\|
            \mathbf{Y}_k-\mathbf{y}_k
        \right\|_2^2,
        & \text{regression}.
    \end{cases}
    \label{eq:Ltask}
\end{equation}
Here $N_b$ is the minibatch size and $k$ indexes samples within the
minibatch. For classification, $\mathbf{Y}_k$ is the one-hot
representation of the ground-truth class label, and $\mathbf{y}_k$ is
the predicted class-probability vector obtained from the logits. For
regression, $\mathbf{Y}_k,\mathbf{y}_k\in\mathbb{R}^{d_y}$ denote the
vectorized target and prediction, respectively, where $d_y$ is the
number of scalar regression targets per sample.

Combining these terms, the training objective is
\begin{equation}
    \mathcal{L}_{\mathrm{total}}
    =
    \mathcal{L}_{\mathrm{task}}
    +
    \sum_{q\in\mathcal{Q}}
    \lambda_q\mathcal{L}_q,
    \label{eq:lossfunction}
\end{equation}
where
$\mathcal{Q}=
\{\mathrm{sim},\mathrm{norm},\mathrm{ortho},
\mathrm{sparsity},\mathrm{aux}\}$
and $\lambda_q$ controls the contribution of each regularization term.
The latent converters ($\mathcal{C}_i$), projector ($\mathcal{P}$), fusion module ($\mathcal{F}$), and task head ($\mathcal{H}$) are
optimized jointly, while the modality-specific encoders remain frozen.

\subsection{Online Query--Response Inference}
\label{sec:online_inference}

During online inference, all model parameters are fixed, and the
training-only redundancy branch is disabled. For each helper $\mathcal{D}_i$, the projector $\mathcal{P}$ generates $A_c^i$ from the current $Z_1$, and the main device encodes the
resulting scalar-level support as a compact index list. Since $A_c^i$ depends on the current $Z_1$, each query is
sample-adaptive and helper-specific. The helper returns the selected latent values in the same order as the
received indices, so the indices need not be transmitted again. The main device places the returned values at the queried indices and reweights them using the corresponding locally retained entries of
$A_c^i$ to form $Z_c^i$. The resulting complementary
representations are then processed by the trained residual fusion
module and task head for prediction. Accordingly, the model-level communication payload consists only of
the support indices and selected helper feature values.

\begin{figure}[!htb]
    \centering
    \includegraphics[width=\columnwidth]{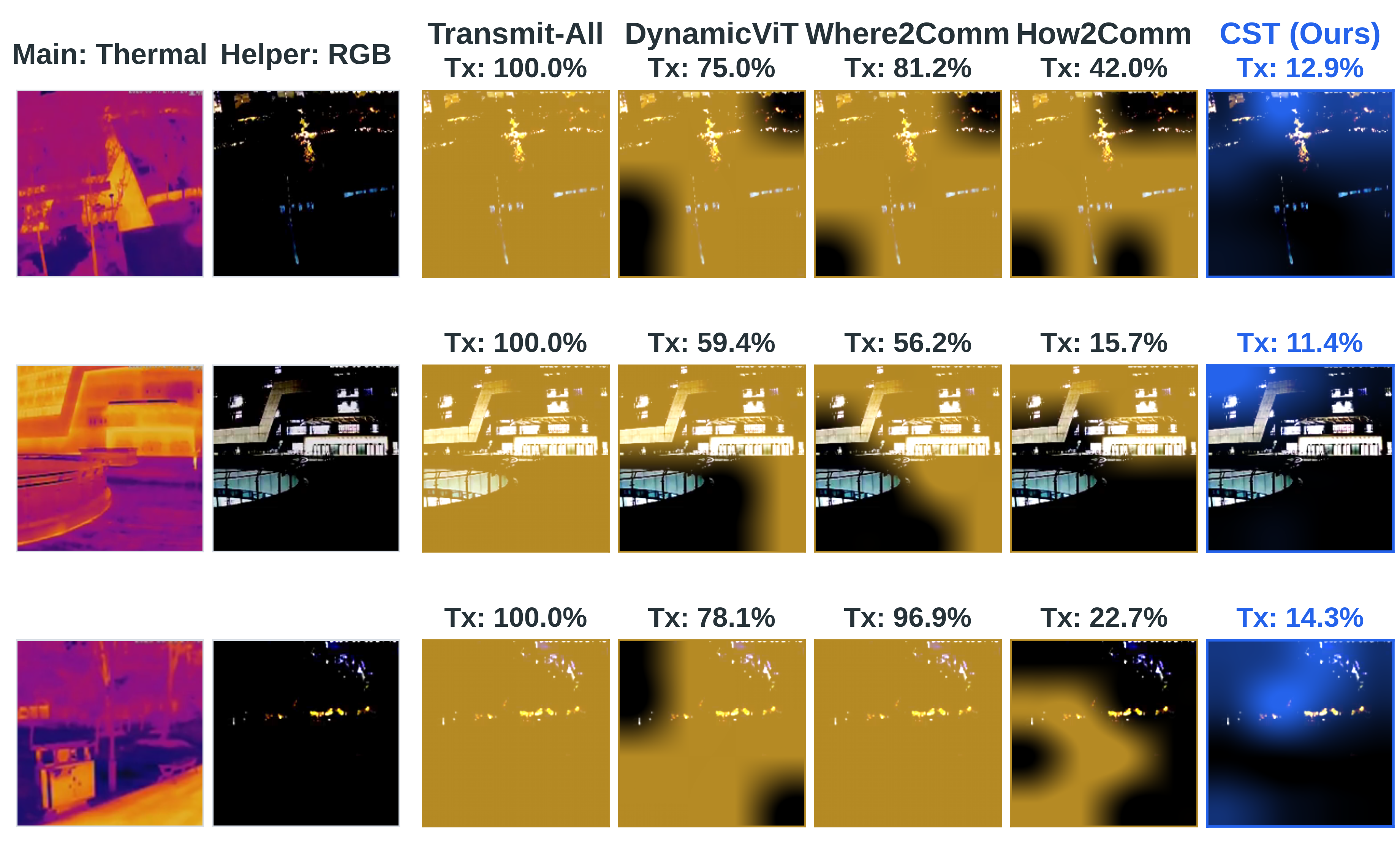}
    \caption{Qualitative RGB-helper selection on DarkAct with Thermal as
    the main modality. Highlighted locations denote retained features.
    CST uses a smaller support concentrated on informative human and scene
    regions.
    }
    \label{fig:case_study_darkact}
\end{figure}

\subsection{Qualitative Case Study}
\label{sec:qualitative_case}

Figure~\ref{fig:case_study_darkact} shows that, compared with the
other methods, whose selected regions are broader, more dispersed,
and often extend into low-contrast background regions, CST retains
no more than 15\% of the RGB helper features while concentrating
its support on key regions of the scene. This comparison suggests that
CST preferentially retains more informative helper evidence that
complements the main representation.

\section{Evaluation}
\label{sec:evaluation}

\subsection{Experimental Setup}
\label{sec:experiment_setup}

\textbf{Datasets and tasks.}
We evaluate CST on three real-world multimodal sensing benchmarks, as
summarized in Table~\ref{tab:dataset_setup}. For each task, we select
the modality with the strongest unimodal validation performance as the
main modality and assign the remaining modalities to helper devices.
The resulting main--helper assignments are determined before CST
training and remain fixed across all compared methods; dynamic main
election is outside the current scope.

\begin{table}[!htb]
\centering
\caption{Datasets, modality assignments, sequence lengths, and task-specific performance metrics.
Main modalities are bold. Metric definitions follow the benchmark papers.}
\label{tab:dataset_setup}

\resizebox{\linewidth}{!}{%
\renewcommand{\arraystretch}{1.25}

\begin{tabular}{@{} l l l l l @{}}
\toprule
\textbf{Dataset}
& \textbf{Task}
& \textbf{Modalities}
& \textbf{Seq. Length ($G_i$)}
& \textbf{Metrics} \\
\midrule

MM-Fi~\cite{mmfi}
& 3D Pose Estimation
& \begin{tabular}{@{}l@{}}
    \textbf{RGB}, Depth, LiDAR, \\
    mmWave, WiFi-CSI
  \end{tabular}
& \begin{tabular}{@{}l@{}}
    32, 180, 32, \\
    32, 32
  \end{tabular}
& \begin{tabular}{@{}l@{}}
    MPJPE, \\
    PA-MPJPE
  \end{tabular} \\
\midrule

CUHK-S~\cite{cuhks}
& Action Recognition
& \begin{tabular}{@{}l@{}}
    \textbf{Skeleton}, Depth/Color, \\
    IR, IMU, Radar
  \end{tabular}
& \begin{tabular}{@{}l@{}}
    32, 196, \\
    196, 128, 20
  \end{tabular}
& \begin{tabular}{@{}l@{}}
    Acc., F1, \\
    Prec., Rec.
  \end{tabular} \\
\midrule

DarkAct~\cite{darkact}
& Action Recognition
& \textbf{Thermal}, RGB
& 32, 32
& Accuracy \\

\bottomrule
\end{tabular}%
}
\end{table}

\textbf{Baselines.}
We use Main-Only and Transmit-All as references. Communication-efficient
baselines include BottleNet++~\cite{bottlenetpp}, VIB-Dyn~\cite{vibdyn},
VDDIB-SR~\cite{vddibsr}, SINFONY~\cite{sinfony}, U-DeepSC-FSM~\cite{udeepfsm}, DT-JSCC~\cite{dtjscc}, DynamicViT~\cite{dynamicvit},
Where2Comm~\cite{where2comm}, and How2Comm~\cite{how2comm}.
All methods use identical modality assignments, data splits, frozen
encoder checkpoints, and encoder-token interfaces of shape $G_i\times512$, where $G_i$ is the modality-specific latent sequence length summarized in Table~\ref{tab:dataset_setup}. They share the same fusion and task-head
architectures, but each method is independently initialized and
trained. Therefore, fusion and task-head weights are not shared. CST's
latent converter and each baseline's adapter, compressor, or decoder
belong to the corresponding method-specific communication module design.

\textbf{Testbed Implementation.}
All models are implemented in PyTorch and trained on an NVIDIA RTX
A5000 GPU. Subsequently, they are deployed on a testbed comprising up to five NVIDIA Jetson Orin Nano nodes, with one modality assigned to each active node. Linux Traffic Control configures the shared aggregate bandwidth, RTT, and jitter.

\subsection{Overall Performance and System Efficiency}

\subsubsection{Task Performance and Feature Transmission Rate}
\label{sec:performance_evaluation}

\begin{table*}[t]
\centering
\caption{Task performance and transmission rate on three
benchmarks. Tx reports the transmitted representation size as a percentage of the
complete helper representations. MM-Fi errors are in millimeters; CUHK-S F1,
precision, and recall are macro-averaged. Actual bidirectional traffic for representative methods measured on
the testbed is reported separately in Table~\ref{tab:protocol_bytes}. $\uparrow$ and $\downarrow$ indicate whether higher or lower values are preferred, respectively. Best observed values are bold.}
\label{tab:overall_performance}

\vspace{-2pt} 
\scriptsize 
\renewcommand{\arraystretch}{0.90} 
\setlength{\tabcolsep}{5.5pt} 

\begin{tabular}{@{} l | ccc | ccccc | cc @{}}
\toprule
\multirow{2}{*}{\textbf{Method}}
& \multicolumn{3}{c|}{\textbf{MM-Fi}}
& \multicolumn{5}{c|}{\textbf{CUHK-S}}
& \multicolumn{2}{c}{\textbf{DarkAct}} \\
\cmidrule(lr){2-4}
\cmidrule(lr){5-9}
\cmidrule(l){10-11}
& MPJPE $\downarrow$
& PA-MPJPE $\downarrow$
& Tx (\%) $\downarrow$
& Acc. (\%) $\uparrow$
& F1 (\%) $\uparrow$
& Prec. (\%) $\uparrow$
& Rec. (\%) $\uparrow$
& Tx (\%) $\downarrow$
& Acc. (\%) $\uparrow$
& Tx (\%) $\downarrow$ \\
\midrule

Main-Only
& 119.20 & 68.94 & --
& 66.28 & 63.43 & 65.50 & 63.90 & --
& 64.55 & -- \\

Transmit-All
& 87.35 & 37.06 & 100.00
& 67.11 & 63.99 & 65.18 & 64.83 & 100.00
& 67.84 & 100.00 \\

\midrule

BottleNet++~\cite{bottlenetpp}
& 86.13 & 38.71 & 50.00
& 68.29 & 66.78 & 69.96 & 67.74 & 50.00
& 66.45 & 50.00 \\

VIB-Dyn~\cite{vibdyn}
& 87.72 & 44.28 & 30.66
& 72.82 & 68.52 & 69.69
& \textbf{69.90} & 30.86
& 66.35 & 30.66 \\

VDDIB-SR~\cite{vddibsr}
& 86.39 & 38.76 & 31.21
& 70.81 & 67.29 & 68.64 & 68.89 & 30.82
& 66.45 & 45.34 \\

U-DeepSC-FSM~\cite{udeepfsm}
& 92.84 & 41.51 & 46.99
& 70.31 & 66.85 & 70.04 & 66.12 & 40.04
& 63.88 & 55.54 \\

SINFONY~\cite{sinfony}
& 82.35 & 36.55 & 50.00
& 70.81 & 67.03 & 67.95 & 68.05 & 50.00
& 67.01 & 50.00 \\

DT-JSCC~\cite{dtjscc}
& 84.76 & 37.02 & 50.00
& 68.96 & 64.98 & 70.44 & 65.16 & 50.00
& 66.83 & 50.00 \\

DynamicViT~\cite{dynamicvit}
& 84.42 & 36.89 & 44.08
& 70.47 & 65.31 & 68.85 & 66.75 & 44.47
& 66.32 & 56.03 \\

Where2Comm~\cite{where2comm}
& 90.49 & 40.02 & 48.75
& 70.81 & 67.86 & 70.54 & 68.29 & 44.89
& 66.64 & 53.00 \\

How2Comm~\cite{how2comm}
& 83.39 & \textbf{36.22} & 45.67
& 68.96 & 64.81 & 65.61 & 66.36 & 39.42
& 67.04 & 31.20 \\

\midrule

\textbf{CST (Ours)}
& \textbf{78.49} & 36.35 & \textbf{11.46}
& \textbf{72.99} & \textbf{68.59} & \textbf{70.81}
& 69.10 & \textbf{8.66}
& \textbf{67.87} & \textbf{14.18} \\

\bottomrule
\end{tabular}
\vspace{-6pt}
\end{table*}

Table~\ref{tab:overall_performance} compares task performance and
feature transmission across the three sensing benchmarks.  At the evaluated operating points, CST achieves the lowest transmission rate in every setting while ranking first or second on all reported task
metrics. On average, CST transmits only 11.43\% of the available helper feature values. These results indicate that CST's communication reduction is not obtained at the expense of task performance.

On MM-Fi, CST achieves the lowest MPJPE while remaining within
0.13\,mm of the best PA-MPJPE, using only 11.46\% of the helper
features. Relative to Transmit-All, it reduces MPJPE by 10.1\% and
helper-feature transmission by 88.54\%. How2Comm achieves the lowest PA-MPJPE, but transmits nearly four times
as many helper feature values as CST. Therefore, CST offers a more favorable balance between pose accuracy and communication cost.

The advantage is also evident on CUHK-S, where CST achieves the highest
accuracy, Macro-F1, and Macro-Precision, together with the
second-highest Macro-Recall. These results are obtained with an 8.66\%
transmission rate. Compared with VIB-Dyn, which provides the strongest
competing classification results, CST achieves slightly higher accuracy and Macro-F1 while reducing Tx
by 71.9\%. Its Macro-Recall is only 0.80 percentage points below the best result. CST
also outperforms Transmit-All across all four classification metrics
despite communicating less than one tenth of the complete helper
representations.

On DarkAct, CST matches Transmit-All within 0.03 percentage points
while transmitting only 14.18\% of the complete helper representation, corresponding to an 85.82\% reduction
in helper feature-value transmission. This result indicates that CST
preserves the task utility of the RGB helper without transmitting its
complete representation.

\subsubsection{Protocol Traffic and Bandwidth Sensitivity}
We also measure actual bidirectional communication traffic and examine
how the resulting communication volume affects end-to-end latency
under varying bandwidth limits. All end-to-end measurements include both
communication directions and all method-specific computation and processing costs.

\begin{table}[!htb]
\centering
\caption{Mean bidirectional traffic (kilobytes per sample) on
the testbed, including requests, responses, framing, and metadata.}
\label{tab:protocol_bytes}

\scriptsize 
\setlength{\tabcolsep}{14pt}
\renewcommand{\arraystretch}{1.08}

\begin{tabular}{@{}lccc@{}}
\toprule
Method
& MM-Fi
& CUHK-S
& DarkAct \\
\midrule
Transmit-All & 565.59 & 1106.30 & 65.65 \\
BottleNet++  & 282.97 & 553.33  & 32.84 \\
VIB-Dyn      & 173.67 & 341.65  & 20.17 \\
VDDIB-SR     & 172.68 & 341.49  & 29.97 \\
SINFONY      & 282.97 & 553.33  & 32.84 \\
DynamicViT   & 189.91 & 482.26  & 36.05 \\
Where2Comm   & 227.18 & 488.82  & 34.81 \\
How2Comm     & 254.14 & 478.83  & 22.77 \\
\textbf{CST} & \textbf{79.92} & \textbf{186.94} & \textbf{11.16} \\
\bottomrule
\end{tabular}
\end{table}

\textbf{Protocol traffic.}
Table~\ref{tab:protocol_bytes} reports the actual bidirectional
communication traffic. Relative to Transmit-All, CST reduces bidirectional traffic by
83.0--85.9\% across the three datasets. Compared with the next-lowest method shown on each dataset, CST further reduces traffic by 44.7--53.7\%. These results show that the extra overhead involved in sending support indices
and packaging the request and response messages is negligible compared to the reduction in transmitted helper features.

\begin{figure}[!htb]
    \centering
    \includegraphics[width=\columnwidth]
    {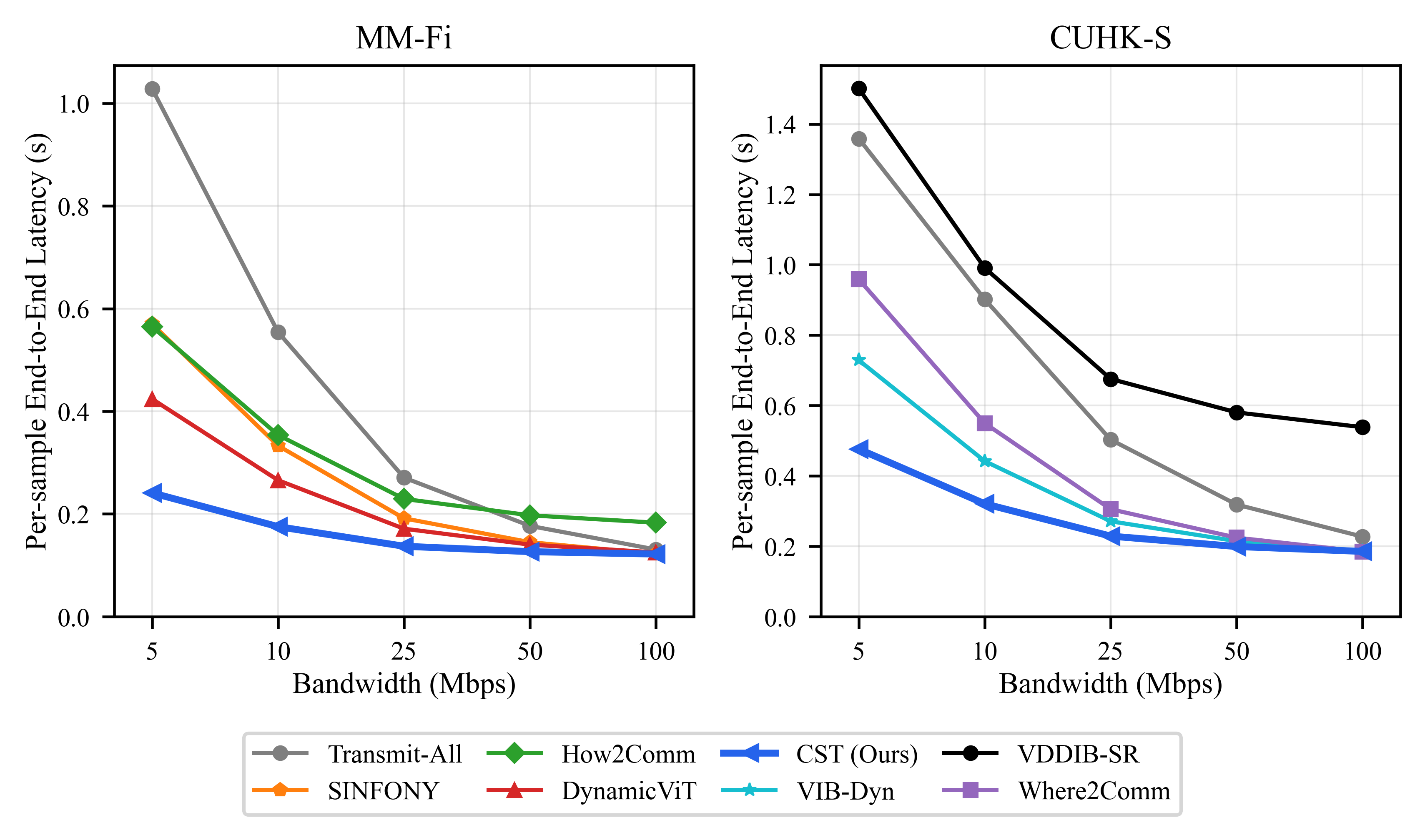}
    \caption{Mean per-sample end-to-end latency, computed as the mean
    complete-batch latency divided by the batch size of 32, versus shared
    aggregate bandwidth on MM-Fi and CUHK-S. RTT and jitter are configured to
    30\,ms and 5\,ms, respectively.}
    \label{fig:bandwidth_latency}
\end{figure}

\textbf{Bandwidth sensitivity.}
Figure~\ref{fig:bandwidth_latency} shows how these traffic reductions
translate into end-to-end latency as the bandwidth limit increases from
5 to 100\,Mbps. At 5\,Mbps, CST achieves per-sample latencies of
approximately 0.24\,s on MM-Fi and 0.48\,s on CUHK-S, corresponding to
$4.27\times$ and $2.85\times$ speedups over Transmit-All. Compared
with the fastest plotted baseline on each dataset, CST reduces latency
by 43.1\% on MM-Fi and 34.7\% on CUHK-S.

Prior collaborative-inference studies have evaluated shared links over
similar bandwidth ranges~\cite{janus,jointconfig2020}. CST remains
15.6--34.0\% faster than the fastest plotted baseline at
10--25\,Mbps, showing that its advantage is not limited to the lowest
bandwidth setting. As bandwidth increases, transmission delay contributes less to the
end-to-end latency, and the gaps among methods consequently narrow. Nevertheless, at 100\,Mbps, CST still records the lowest observed latency, although
the margin over the fastest baseline is small.

\subsubsection{Sensitivity to Query--Response RTT}
Because CST completes a request--response exchange before fusion, RTT
introduces an additional source of communication delay. We therefore
vary the configured RTT while fixing bandwidth, jitter, and batch size
to evaluate whether CST retains its end-to-end latency advantage as
this delay increases.

\begin{table}[!htb]
\centering
\caption{Mean and p95 complete-batch end-to-end latency on MM-Fi under varying configured RTTs. The bandwidth is 10\,Mbps, the jitter target is 5\,ms, and the batch size is 16. Values are in seconds and computed over 20 measured batches.}
\label{tab:rtt_latency}

\footnotesize 
\setlength{\tabcolsep}{3.8pt} 
\renewcommand{\arraystretch}{1.08}

\begin{tabular}{@{}lcccccccc@{}}
\toprule
Method
& \multicolumn{2}{c}{10\,ms}
& \multicolumn{2}{c}{30\,ms}
& \multicolumn{2}{c}{50\,ms}
& \multicolumn{2}{c}{100\,ms} \\
\cmidrule(lr){2-3}
\cmidrule(lr){4-5}
\cmidrule(lr){6-7}
\cmidrule(lr){8-9}
& Mean & p95
& Mean & p95
& Mean & p95
& Mean & p95 \\
\midrule
Transmit-All
& 8.95 & 9.31
& 8.76 & 8.78
& 8.80 & 8.83
& 8.88 & 8.92 \\

SINFONY
& 5.03 & 5.37
& 4.99 & 4.99
& 4.99 & 5.01
& 5.12 & 5.20 \\

DynamicViT
& 3.78 & 4.13
& 3.89 & 4.27
& 3.87 & 4.31
& 3.99 & 4.39 \\

How2Comm
& 5.31 & 5.71
& 5.36 & 5.75
& 5.41 & 5.78
& 5.62 & 6.01 \\

\textbf{CST}
& \textbf{2.86} & \textbf{2.89}
& \textbf{3.05} & \textbf{3.10}
& \textbf{2.97} & \textbf{3.00}
& \textbf{3.26} & \textbf{3.29} \\
\bottomrule
\end{tabular}
\end{table}

Table~\ref{tab:rtt_latency} reports the mean and 95th-percentile (p95) complete-batch end-to-end latency at configured RTTs from 10 to 100\,ms. CST achieves the lowest mean and p95 latency at every evaluated RTT. Even at 100\,ms RTT, its mean and p95 latencies remain 18.3\% and 25.1\%
lower than DynamicViT, respectively. These results indicate that
increasing RTT does not substantially reduce CST's end-to-end latency
advantage over the evaluated range.

\subsubsection{Scalability with Batch Size}
\begin{figure}[!htb]
    \centering
    \includegraphics[width=\columnwidth]
    {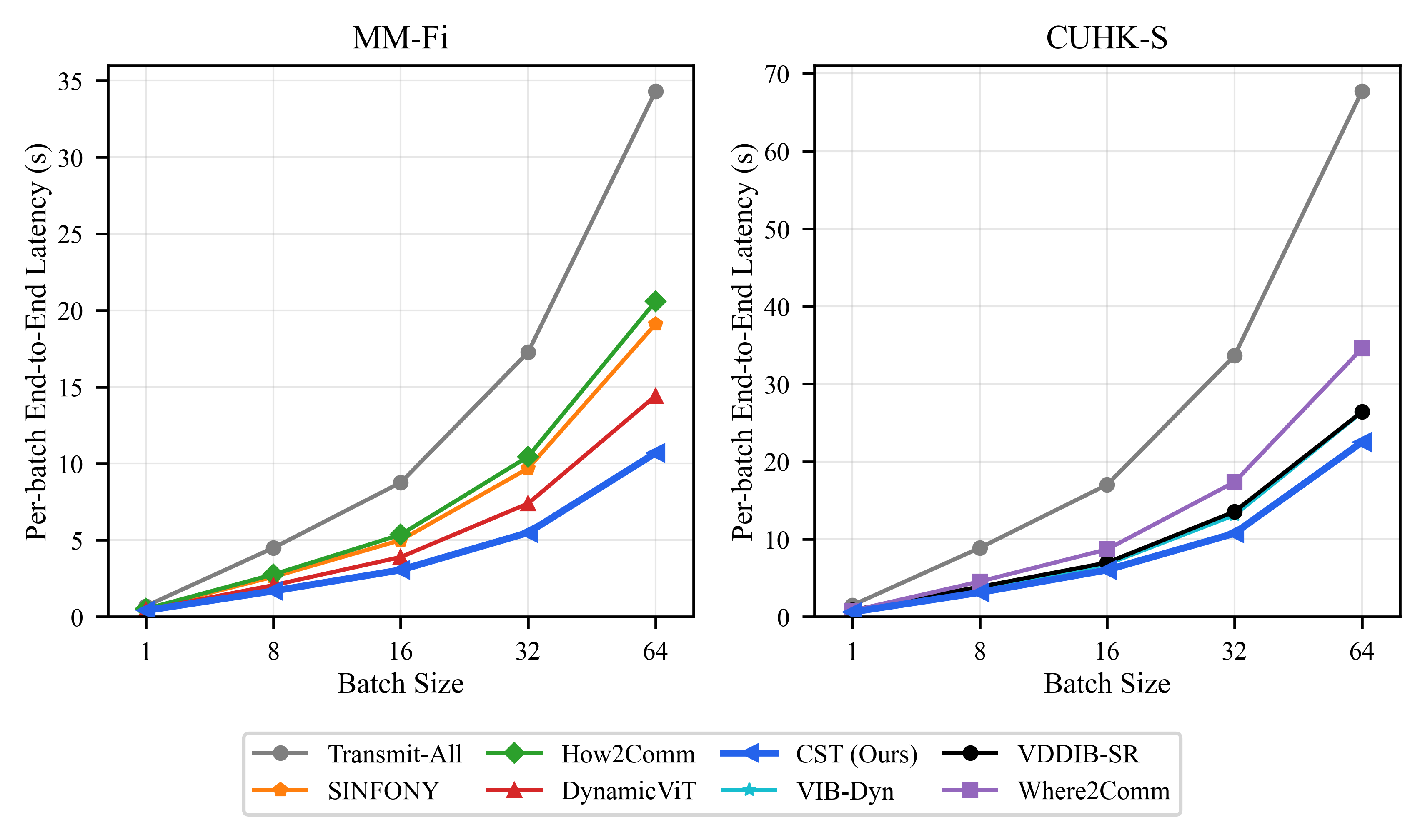}
    \caption{Mean complete-batch end-to-end latency
    versus batch size on MM-Fi and CUHK-S. The
    shared aggregate bandwidth is 10\,Mbps, with
    configured RTT and jitter targets of 30\,ms and
    5\,ms, respectively.}
    \label{fig:batch_latency}
\end{figure}

Figure~\ref{fig:batch_latency} evaluates batch sizes from 1 to 64 under
a fixed shared aggregate bandwidth of 10\,Mbps. At
small batch sizes, the transmitted payload is small and the latency
differences among the fastest methods remain small. As the batch size
increases, however, the communication cost accumulates across samples,
and the curves separate according to their per-sample transmission
requirements. CST maintains the lowest latency among the plotted
methods across the complete batch-size range on both datasets.

At a batch size of 64, CST completes the full MM-Fi and CUHK-S
workloads in 10.69\,s and 22.51\,s, respectively, yielding
$3.21\times$ and $3.01\times$ speedups over Transmit-All. Its
latency also remains 25.9\% below DynamicViT on MM-Fi and 14.7\% below
VIB-Dyn on CUHK-S, the fastest plotted baselines. Together, the bandwidth and batch-size sweeps demonstrate that CST's
lower per-sample transmission volume results in consistently lower
end-to-end latency under both low-bandwidth settings and large
inference batches.

\subsection{Mechanism Analysis and Insights}
\subsubsection{Ablation Study}
\label{sec:ablation}

Table~\ref{tab:cst_ablation} presents an ablation analysis of two
design aspects: the necessity of the redundancy anchor
$A_r$ for compact complementary selection and the contribution of
each auxiliary objective to task performance and communication
selectivity.

\begin{table}[!htb]
\centering
\caption{CUHK-S ablation results. Direct-$A_c$ uses only
$\mathcal L_{\mathrm{task}}$ and $\mathcal L_{\mathrm{sparsity}}$;
the second variant retains $\mathcal L_{\mathrm{aux}}$ while removing
the redundancy branch. All values are percentages, and best values are bold.}
\label{tab:cst_ablation}

\vspace{-2pt}
\scriptsize
\setlength{\tabcolsep}{7.5pt} 
\renewcommand{\arraystretch}{1.02} 

\begin{tabular}{@{}lccccc@{}}
\toprule
Variant
& Acc. $\uparrow$
& F1 $\uparrow$
& Prec. $\uparrow$
& Rec. $\uparrow$
& Tx (\%) $\downarrow$ \\
\midrule

\textbf{Full CST}
& \textbf{72.99}
& \textbf{68.59}
& \textbf{70.81}
& \textbf{69.10}
& \textbf{8.66} \\

Direct-$A_c$
& 72.32
& 67.90
& 70.79
& 68.59
& 51.82 \\

Direct-$A_c$+$\mathcal{L}_{\mathrm{aux}}$
& 72.48
& 68.27
& 70.72
& 68.90
& 50.79 \\

\midrule

w/o $\mathcal{L}_{\mathrm{sim}}$
& 72.31
& 67.57
& 70.42
& 67.85
& 42.34 \\

w/o $\mathcal{L}_{\mathrm{ortho}}$
& 72.32
& 68.03
& 70.72
& 69.04
& 36.62 \\

w/o $\mathcal{L}_{\mathrm{aux}}$
& 72.82
& 68.30
& 70.17
& 68.91
& 37.26 \\

w/o $\mathcal{L}_{\mathrm{sparsity}}$
& 72.65
& 68.53
& 69.74
& 68.56
& 56.35 \\

\bottomrule
\end{tabular}
\vspace{-6pt}
\end{table}

\textbf{Necessity of the redundancy branch.}
We remove $A_r$ and all associated objectives to construct two reduced variants. Direct-$A_c$ uses only $\mathcal L_{\mathrm{task}}$ and $\mathcal L_{\mathrm{sparsity}}$, while Direct-$A_c$+$\mathcal L_{\mathrm{aux}}$ additionally retains cross-helper regularization. Both variants maintain similar task performance but yield Tx values of 51.82\% and 50.79\%, compared with 8.66\% for full CST. This large Tx gap highlights the essential role of the $A_r$-based redundancy guidance in learning a compact complementary support. The slight Tx reduction after adding $\mathcal L_{\mathrm{aux}}$ indicates that cross-helper regularization helps reduce some overlap among helpers, but does not identify helper information already covered by the main representation. 

\textbf{Contribution of individual objectives.}
Removing $\mathcal L_{\mathrm{sparsity}}$,
$\mathcal L_{\mathrm{sim}}$, $\mathcal L_{\mathrm{ortho}}$, or
$\mathcal L_{\mathrm{aux}}$ raises Tx to 56.35\%, 42.34\%, 36.62\%,
and 37.26\%, respectively, while task performance changes only
marginally. Taken together, these results show that the auxiliary
objectives primarily improve communication selectivity:
$\mathcal L_{\mathrm{sparsity}}$ promotes compact support,
$\mathcal L_{\mathrm{sim}}$ and $\mathcal L_{\mathrm{ortho}}$
establish and separate the main-covered reference, and
$\mathcal L_{\mathrm{aux}}$ reduces cross-helper overlap.

\subsubsection{Post-hoc Partial Information Decomposition Analysis}
\label{sec:posthoc_pid}

To assess whether reduced transmission preserves task-relevant helper
information, we apply Flow-PID~\cite{flowpid} to the DarkAct dataset after task
training. For all methods, Thermal is the fixed main representation,
RGB is the helper representation delivered for fusion, and the action
label is the target. Flow-PID first maps the main representation, helper representation,
and target to approximately Gaussian pairwise marginals, and then uses
its Thin-PID solver to estimate the information components. We report
helper-unique information $U_{\mathrm{H}}$, synergy $S$, and
$C=U_{\mathrm{H}}+S$, where $C$ quantifies the task-relevant information
contributed by the delivered RGB representation beyond the fixed main
Thermal representation in terms of information-theoretic bits. Tx is included for communication-cost comparison. 

\begin{table}[!htb]
\centering
\caption{Post-hoc Flow-PID results on DarkAct, reported in bits as
mean $\pm$ standard deviation over three fixed Flow initializations. Upward and downward arrows indicate that higher and lower values are preferred, respectively. The largest mean $C$ and lowest
Tx among communication-efficient methods are bold.}
\label{tab:darkact_pid}

\renewcommand{\arraystretch}{1.08}
\setlength{\tabcolsep}{3.0pt}
\resizebox{\columnwidth}{!}{%
\begin{tabular}{@{}lcccc@{}}
\toprule
Method
& $U_{\mathrm{H}}$
& $S$
& $C=U_{\mathrm{H}}+S \uparrow$
& Tx (\%) $\downarrow$ \\
\midrule

Transmit-All
& $1.783 \pm 0.078$
& $1.464 \pm 0.057$
& $3.247 \pm 0.125$
& $100.00$ \\
\midrule

BottleNet++
& $1.602 \pm 0.044$
& $1.406 \pm 0.046$
& $3.008 \pm 0.078$
& $50.00$ \\

VIB-Dyn
& $1.576 \pm 0.047$
& $1.403 \pm 0.047$
& $2.979 \pm 0.078$
& $30.66$ \\

SINFONY
& $1.651 \pm 0.111$
& $1.375 \pm 0.078$
& $3.026 \pm 0.166$
& $50.00$ \\

DT-JSCC
& $1.659 \pm 0.133$
& $1.363 \pm 0.077$
& $3.023 \pm 0.197$
& $50.00$ \\

DynamicViT
& $1.712 \pm 0.154$
& $1.290 \pm 0.088$
& $3.002 \pm 0.216$
& $56.03$ \\

Where2Comm
& $1.816 \pm 0.133$
& $1.298 \pm 0.014$
& $3.113 \pm 0.124$
& $53.00$ \\

How2Comm
& $1.787 \pm 0.154$
& $1.370 \pm 0.060$
& $\mathbf{3.158 \pm 0.200}$
& $31.20$ \\

\textbf{CST-$Z_c$ (Ours)}
& $1.640 \pm 0.095$
& $1.519 \pm 0.029$
& $\mathbf{3.158 \pm 0.116}$
& $\mathbf{14.18}$ \\

\bottomrule
\end{tabular}%
}
\end{table}

Table~\ref{tab:darkact_pid} shows that CST retains
$C=3.158$ bits with only 14.18\% Tx, preserving 97.3\% of the conditional task information provided by the
Transmit-All helper representation while reducing
feature-value Tx by 85.82\%. At the reported precision, CST and How2Comm yield the same $C$, while
CST uses 54.6\% lower Tx. Note that $C$ is not a monotonic predictor of accuracy. It indicates information retention rather than direct performance equivalence.

\section{Related Work}

\subsection{Collaborative Inference and Selective Communication}

Collaborative inference improves edge perception by exchanging
representations among distributed sensing devices, but dense feature
exchange can impose substantial communication overhead.
Communication-graph and partner-selection methods reduce the number of
participating agents~\cite{when2com,which2comm}. Spatial confidence
and spatial--channel requests further restrict communication to
informative regions or channels~\cite{where2comm,how2comm}.
CodeFilling jointly optimizes coded message representations and
information-demand-aware selection~\cite{codefilling}. Multi-resolution collaborative perception further combines selective-region communication with collaborative reconstruction
~\cite{umc2023}.

These methods primarily determine which agents, regions, channels, or
coded messages should be exchanged. CST instead uses the current main
representation to generate helper-specific scalar retrieval supports,
so helpers return only the requested latent values, while a
training-time redundancy reference discourages the retrieval of
main-covered semantics.

\subsection{Task-Oriented Feature Compression and Transmission}
Split computing and pipelined inference optimize single-model execution
across device--edge/cloud resources~\cite{splitcomputingearlyexiting,dads2019}.
Systems adapt partitioning to runtime conditions~\cite{deepdecision,spinn} or
distribute execution across heterogeneous edge
resources~\cite{edgeflow2022,coscheduling2025}. CST instead considers
independently encoded, synchronized modalities and determines which
helper values remain useful given each sample's main
representation.

Compression and information-bottleneck methods construct lower-rate
source messages~\cite{bottlenetpp,vibdyn,vddibsr}, while semantic and
intermediate-feature methods optimize compact task-oriented
representations~\cite{sinfony,dtjscc,frankensplit2024}.
Input-adaptive pruning ranks content by source-local
importance~\cite{dynamicvit}, while retransmission requests additional
encoded content~\cite{vddibsr}. CST instead generates helper-specific
supports at the main device and directly retrieves the corresponding
latent values.

\subsection{Multimodal Redundancy and Information Decomposition}

Different modalities may provide task-relevant information that is
shared across views, specific to one modality, or useful only through
their combination. Multi-view representation learning commonly
exploits shared information under the Multiview Redundancy
Assumption~\cite{tsai2021selfsupervisedlearningmultiviewperspective},
while recent methods further characterize modality-unique and
synergistic interactions~\cite{FactorCL,CoMM,
liang2023quantifyingmodelingmultimodal}. Partial Information
Decomposition provides a formal framework for distinguishing
redundant, unique, and synergistic contributions to a target
variable~\cite{demopositionMI,quantifyuniqueinfo}. Existing work
primarily applies these concepts to representation learning, fusion,
or post-hoc interaction analysis~\cite{liang2023quantifyingmodelingmultimodal, tsai2021selfsupervisedlearningmultiviewperspective}. CST instead uses this perspective to guide communication before dense
helper features are transmitted, without explicitly estimating PID
components during training.

\section{Conclusion}
\label{sec:conclusion}
This paper introduced CST, a main-directed sparse retrieval framework
that learns sample-adaptive, helper-specific supports using a
training-time redundancy reference. Across three multimodal
benchmarks, CST achieves best or near-best task performance among the evaluated
methods while transmitting at most 14.18\% of helper feature values. A five-node NVIDIA Jetson Orin Nano testbed shows up to a
$4.27\times$ end-to-end speedup under a shared aggregate bandwidth of
5\,Mbps. Post-hoc Flow-PID analysis further indicates that the sparse RGB
representation transmitted by CST preserves 97.3\% of the conditional
task information provided by the Transmit-All helper representation. Overall, these results demonstrate that CST enables accurate, communication-efficient, and low-latency multimodal edge inference.

\bibliographystyle{IEEEtran}
\bibliography{references}

\end{document}